\documentclass{article}

\usepackage{microtype}
\usepackage{graphicx}
\usepackage{subfigure}
\usepackage{booktabs} 
\usepackage{bm}
\usepackage{amsmath}
\usepackage{amssymb}
\usepackage{amsthm}

\RequirePackage{algorithm}
\RequirePackage{algorithmic}

\theoremstyle{plain}
\newtheorem{theorem}{Theorem}[section]

\newtheorem{limitation}[theorem]{Limitation}
\newtheorem{lemma}[theorem]{Lemma}
\newtheorem{corollary}[theorem]{Corollary}
\theoremstyle{definition}
\newtheorem{definition}[theorem]{Definition}
\newtheorem{assumption}[theorem]{Assumption}
\theoremstyle{remark}

\newcommand{\acrold}{ONO}
\newcommand{\acr}{PHO}

\newcommand{\acrgam}{PHOGAM}

\usepackage{hyperref}

\usepackage[accepted]{icml2023}

\icmltitlerunning{Post-hoc Orthogonalization}

\begin{document}

\twocolumn[
\icmltitle{A New PHO-rmula for Improved Performance of Semi-Structured Networks}



\icmlsetsymbol{equal}{*}

\begin{icmlauthorlist}
\icmlauthor{David R\"ugamer}{lmu,mcml}
\end{icmlauthorlist}

\icmlaffiliation{lmu}{Department of Statistics, LMU Munich, Munich, Germany}
\icmlaffiliation{mcml}{Munich Center for Machine Learning (MCML), Munich, Germany}

\icmlcorrespondingauthor{David R\"ugamer}{david@stat.uni-muenchen.de}

\icmlkeywords{Machine Learning, ICML}

\vskip 0.3in
]



\printAffiliationsAndNotice{} 

\begin{abstract}
Recent advances to combine structured regression models and deep neural networks for better interpretability, more expressiveness, and statistically valid uncertainty quantification demonstrate the versatility of semi-structured neural networks (SSNs). We show that techniques to properly identify the contributions of the different model components in SSNs, however, lead to suboptimal network estimation, slower convergence, and degenerated or erroneous predictions. In order to solve these problems while preserving favorable model properties, we propose a non-invasive post-hoc orthogonalization (PHO) that guarantees identifiability of model components and provides better estimation and prediction quality. Our theoretical findings are supported by numerical experiments, a benchmark comparison as well as a real-world application to COVID-19 infections. 
\end{abstract}

\section{Introduction}

\begin{figure}
    \centering
    \includegraphics[width=0.95\columnwidth]{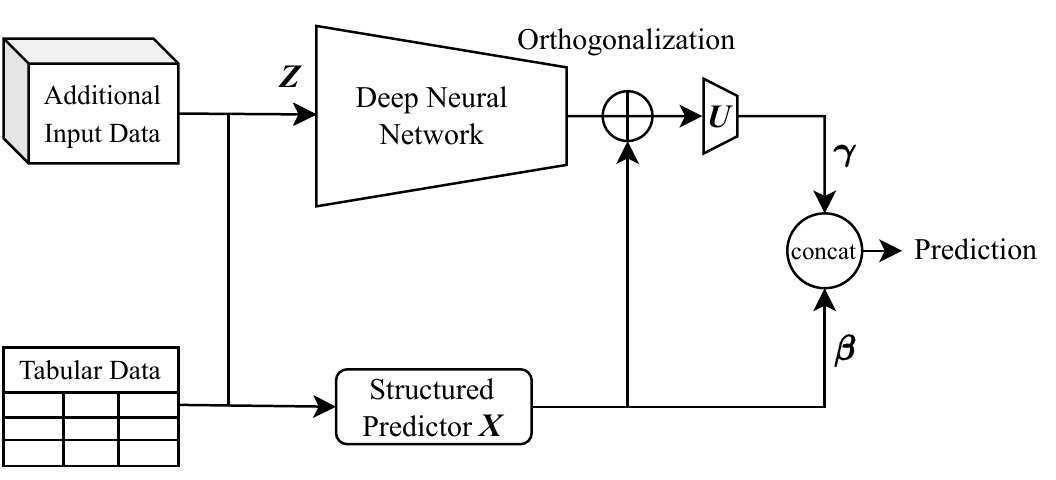}
    \caption{Exemplary SSN learning latent features $\bm{U}\bm{\gamma}$ from tabular and additional input data sources through a DNN and combining these with structured effects $\bm{X}\bm{\beta}$ based on tabular data $\bm{X}$. As $\bm{X}$ is contained in both network parts, an orthogonalization (top right) is required to identify the true linear effect $\bm{\beta}$.}
    \label{fig:ssn}
\end{figure}

A linear model in its original form is inherently interpretable due to its structural model space assumption: Given features $\bm{X}$, the expected outcome $\mathbb{E}(\bm{y}|\bm{X})$ of a variable of interest $\bm{y}$ is the linear combination $\bm{X}\bm{\beta}$ of the features $\bm{X}$ and weights $\bm{\beta}$. When fixing all but one of the features $\bm{x}_j$, the change in $\mathbb{E}(\bm{y}|\bm{X})$ can be easily quantified and interpreted by the change in $\bm{x}_j$ multiplied by its weight $\beta_j$ (ceteris paribus). A deep neural network (DNN), on the other hand, can also be seen as a linear combination of features and weights, say $\bm{U}\bm{\gamma}$, where $\bm{U}$ are latent features learned in the penultimate layer of the neural network and $\bm{\gamma}$ are the weights from the connection between the last hidden and the output layer (in this case with linear activation function). As $\bm{U}$ itself is often a non-linear and complex transformation of the actual DNN inputs $\bm{Z}$, the linear combination $\bm{U}\bm{\gamma}$ is not ``structured'' in a way that users can directly relate the effects $\bm{\gamma}$ to the original inputs $\bm{Z}$. While there is a plethora of literature characterizing the learned DNN effects in terms of $\bm{Z}$, we in this work focus on models that combine a structured linear model-type predictor $\bm{\eta}^{str} := \bm{X}\bm{\beta}$ and an unstructured DNN predictor $\bm{\eta}^{unstr} := \bm{U}\bm{\gamma}$ to combine interpretability and predictive performance. The latent features $\bm{U}$ are learned from a set of features $\bm{Z}$ with arbitrary shape (e.g., image tensors) and potentially overlapping with $\bm{X}$. An example of such a model is the semi-structured neural network (SSN). SSNs, exemplary depicted in Figure~\ref{fig:ssn}, assume a late fusion of the structured and unstructured model by combining both model parts additively, i.e.,
\begin{equation} \label{eq:example}
\bm{\eta}^{str} + \bm{\eta}^{unstr} = \bm{X} \bm{\beta} + \bm{U} \bm{\gamma}.
\end{equation}

This simple fusion has some attractive properties. \citet{dorigatti}, e.g., promote SSNs in the medical application context where researchers require some parts of the network to be a white box and derive statistically valid uncertainty quantification for $\bm{\eta}^{str}$. We will motivate our proposed approach in this paper using SSNs, but later results also generalize to more complex network structures. 

\paragraph{Why make structured model parts explicit?} As DNNs can be designed very flexibly and deep architectures often come with some form of universal approximation guarantee, DNNs can also capture linear effects. This raises the question as to why the structured part needs to be made explicit in models such as the SSN. The reason is the same as for the usage and success of residual connections \citep{he2016deep}: while DNNs can represent increasingly more complex and non-linear functions with greater depth, it also gets increasingly more difficult to learn the identity function of some input for these networks. By concatenating the outputs of (parts of) a DNN with the inputs of this DNN, the identity map can be made explicit and thereby allows the combined network to better learn linear effects of the inputs. 

\paragraph{Identifiability} In late semi-structured fusion models such as the SSN, the interpretability of the structured part can not be preserved without additional constraints \citep[see, e.g.,][]{baumann.2020, fritz2021combining, sddr}. This is due to an identifiability problem. The DNN can also capture linear effects -- to some extent -- but it is unclear to the analyst, how much additional linearity in addition to $\bm{\eta}^{str}$ has been learned by the DNN in $\bm{\eta}^{unstr}$. An obvious solution is to restrict the DNN in such a way that only the structured model part can learn structured effects and the unstructured deep network part captures the remaining variation while preserving the predictive performance of the DNN. While current literature takes this route and suggests that it is necessary to constrain the unstructured model part to ensure the interpretability of $\bm{\eta}^{str}$, restricting the network might potentially deteriorate model performance. 

\paragraph{Our Contributions}

In this work, we provide new insights into approaches that combine structured and unstructured predictors and answer the following open questions:
\vspace*{-0.2cm}
\begin{itemize}
    \item Why and when do constrained SSNs work?
    \vspace*{-0.1cm}
    \item What are the limitations of semi-structured approaches with enforced identifiability?
    \vspace*{-0.1cm}
    \item How can existing limitations be overcome?
    \vspace*{-0.1cm}
    \item How can SSN approaches be generalized?
\end{itemize}
\vspace*{-0.2cm}
Our investigations allow advancing research in the direction of SSNs and interpretability in neural networks in general. Our proposed solutions to existing limitations are generic and applicable wherever a structured linear assumption is made, thereby also addressing use cases beyond SSNs. More specifically, we propose a ``non-invasive'' method to obtain valid interpretability post-model fitting and show both theoretically and in numerical experiments that this approach is superior to existing methods. 

\section{Related Literature} \label{sec:rellit}

Methods that represent or learn structured model parts in a neural network are discussed in the literature in many different aspects. \citet{De.2011}, e.g., discuss how generalized additive models (GAMs) can be framed as a neural network, laying the foundation to combine GAMs with other network parts within one large neural network. In particular, this allows scaling GAMs to large amounts of observations or features \citep[see, e.g.,]{rugamer2023factorized}. In contrast, neural additive models \citep[NAMs;][]{agarwal2021neural} combine neural networks and additive models by learning the basis representations of additive effects using a neural network. 
Generalized linear models and neural network predictors have been combined by \citet{Tran.2018} as well as \citet{Hubin.2018}, and used under the name \emph{wide \& deep learning} for product recommendation by \citet{Cheng.2016}. \citet{Poelsterl.2020} and \citet{kopper2021semi,kopper2021} use the idea of combining structured and unstructured predictors to learn a survival model. A combination of state space models and neural
networks has, e.g., been proposed by~\citet{amoura2011state}. Another strain of literature combines transformation models and DNNs \citep{baumann.2020, sick2021deep} in a semi-structured manner to, e.g., facilitate the combination of structured predictors and complex DNNs for ordinal regression  \citep{KOOK2022108263}. 
Next to different model combinations, properties of semi-structured approaches such as the previously mentioned identifiability \citep{sddr} and also uncertainty quantification of SSNs \citep{dorigatti} have been discussed. 

We will focus on the aspect of identifiability in this paper and point out limitations in existing approaches. Our suggested solutions guarantee correct interpretability, are applicable to most of the methods and models discussed in the literature, and are efficient to compute without modifying the SSN.

\section{Identifiability and Limitations} \label{sec:idenlim}


\citet{sddr} discuss the identifiability problem in the setup of a probabilistic density regression network that learns $K\geq 1$ distribution parameters $\theta_k, k=1,\ldots,K$ using the Maximum Likelihood approach. In their framework, each of the distribution parameters $\theta_k$ is linked to an additive predictor
\begin{equation*}
    \bm{{\eta}}_k = \bm{X}_k\bm{\beta}_k + \bm{U}_k\bm{\gamma}_k.
\end{equation*}
For the purpose of this paper, we simplify this approach and focus on the special case of mean regression. 

\paragraph{Notation}
We assume $n$ observations $\bm{y} \in \mathbb{R}^n$ drawn (conditionally) independent from a distribution $\mathcal{D}_{Y|X,Z}$ based on a fixed set of features $\bm{X} \in \mathbb{R}^{n\times p}$ and a set of features $\bm{Z}$ with arbitrary shape (e.g., image tensors) that are potentially overlapping with $\bm{X}$. As model for the data, we consider an SSN. In the SSN, features $\bm{X}$ are modeled as a structured linear effect with weights $\bm{\beta}\in\mathbb{R}^p$. For the unstructured model part, a DNN with $\bm{Z}$ as input learns latent features $\bm{U}\in\mathbb{R}^{n\times q}$. These latent features are weighted in a final layer with weights $\bm{\gamma}\in\mathbb{R}^{q}$ and then added to the structured model part (cf.~Fig.~\ref{fig:ssn}). Put together, the SSN learns 
\begin{equation} \label{eq:ssmean}
\mathbb{E}(\bm{y}|\bm{X},\bm{Z}) = \bm{\eta} = \bm{X}\bm{\beta} + \bm{U}\bm{\gamma} 
\end{equation}
and we optimize the network using the mean squared error $\ell(\bm{\eta}) = \frac{1}{2n}||\bm{y}-{\bm{\eta}}||_2^2$. For better readability, we focus on mean regression and not a general distributional regression as in \citet{sddr}, but note that results presented in the following also hold for the distributional case.

\paragraph{Identifiability}
Based on this (simplified) regression approach defined in \eqref{eq:ssmean}, we adapt the definition of \emph{semi-structured identifiability} as follows:

\begin{definition}{\textbf{Semi-structured identifiability}}\label{def:ident}
We say that a semi-structured regression model is identified in its structured model parts if there exists no $\bm\xi \in \mathbb{R}^n \backslash \{\bm{0} \}$ such that
\begin{equation}
    \begin{aligned}
\bm\eta &= \bm\eta^{str} + \bm\eta^{unstr} = (\bm\eta^{str} - \bm\xi) + (\bm\eta^{unstr} + \bm\xi)\\
&= \breve{\bm\eta}^{str} + \breve{\bm\eta}^{unstr} = \breve{\bm\eta},       
    \end{aligned}
\end{equation}
where $\ell(\breve{\bm{\eta}}) = \ell(\bm{\eta})$, i.e., both parameterizations perform equally well and yield the same loss.
\end{definition}
Simply put, the above definition says that the structured predictor is identified, if we cannot find another truly different structured predictor that yields the same loss as the original predictor. In particular, the definition only relates to the structured part as $\bm{\eta}^{unstr}$, a DNN, is in most cases never uniquely identified. The solution to guarantee identifiability in \citet{sddr} proposes an orthogonalization cell that projects the unstructured (DNN) predictor $\bm{U}$ into the orthogonal complement space $\mathcal{X}^\bot$ spanned by the columns space of $\bm{X}$, thereby ensuring identifiability and thus, in turn, meaningful interpretability of the structured effects $\bm{\beta}$. More specifically, if the model's predictor is as in \eqref{eq:ssmean}, the adapted SSN with orthogonalization computes the mean via
\begin{equation} \label{eq:ssmeanoz}
 \mathbb{E}(\bm{y}|\bm{X},\bm{Z}) = \bm{\eta} = \bm{X}\bm{\beta} + \mathcal{P}_X^\bot\bm{U}\bm{\gamma}, 
\end{equation}
where $\mathcal{P}_X^\bot = (\bm{I}_{n\times n} - \mathcal{P}_X) \in \mathbb{R}^{n\times n}$ is the orthogonal projection matrix defined as the difference between the identity matrix $\bm{I}_{n\times n}$ and the projection matrix $\mathcal{P}_X$ onto the space $\mathcal{X}$ spanned by $\bm{X}$.
We will call this procedure ``online orthogonalization'' (\acrold) as the respective constraints and orthogonalization operations are applied during training. 

In the following, we will formalize some of the limitations of the \acrold~approach and then propose solutions addressing these restrictions.  

\subsection{Limitations} \label{sec:limitations}

\citet{sddr} proved that \acrold~yields identifiability of the structured model part and thereby fulfilling the purpose of preserving its interpretability. Before addressing some downsides induced by this approach, we first show another desirable property of their orthogonalization approach not mentioned in their work. 
\begin{lemma}{\textbf{\acrold~Hypothesis space}} \label{lemma:hyp}
Let $n > p$. Then the hypothesis space $\mathcal{H}_{ONO}$ of an SSN with orthogonalization is equivalent to the hypothesis space $\mathcal{H}$ of the same SSN without the orthogonalization constraint.
\end{lemma}
The proof of Lemma~\ref{lemma:hyp} can be found in the Appendix. Contrary to the intuition that the orthogonalization directly entails a constraint in the model space, the previous result shows that an SSN with \acrold~is \emph{a priori} as expressive as the same model without the constraint. \acrold, however, induces a different optimization problem, resulting in slower convergence in practice and a restricted hypothesis space if the $n>p$-assumption of Lemma~\ref{lemma:hyp} is not met. We formalize this in the following:
\begin{limitation}[\citet{sddr}, Remark 2.5] \label{lim:null1}
In the case where $p > n$, the orthogonalized unstructured effect is equal to $\bm{0}$.
\end{limitation}
This is due to the fact that $\mathcal{P}_X^\bot = \bm{0}_{n\times n}$  for $p>n$. Using the same argument, we have the following limitation for mini-batch training:
\begin{limitation} \label{lim:null2}
    In the case where $p < n$ but the batch size $b < p$, the theoretical unstructured effect is different from $\bm{0}$ but is practically estimated as a null-effect due to the mini-batch training.
\end{limitation}
The proof is a simple result of the fact that $\mathcal{P}_X^\bot$ is a $b\times b$ matrix formed for every batch in mini-batch training and by construction, $\mathcal{P}_X^\bot = \bm{0}_{b\times b}$ if $b<p$. This finding also relates to the network's prediction as stated in the following:
\begin{limitation} \label{lim:size}
    Due to the orthogonalization, the prediction of an orthogonalized network depends on the batch size used for prediction.
\end{limitation}
This is due to the fact that the network's prediction is formed through the matrix $\mathcal{P}^\bot_X \in \mathbb{R}^{b\times b}$, implying that every of the $b$ observations during prediction influences the $b-1$ other observations (whereas in many other networks, the predictions are independent of each other for fixed network weights). As a consequence, an SSN network with \acrold~will have an additional source of prediction error $\sigma_E^2$, irrespective of its optimization quality when the batch size $b$ at test time is small. Let $\bm{X}^\ast$ and $\bm{Z}^\ast$ be the test data sets fed into the structured and unstructured model part, respectively, and let $\hat{\bm{U}}^\ast = \hat{\bm{U}}(\bm{Z}^\ast)$ be the penultimate layer's features on this test set. We can quantify the behavior of this error w.r.t.~$b$ exemplarily using $p=q=1$ as follows:
\begin{lemma}[\textbf{Error rate ONO prediction}] \label{lemma:error}
Assume that the latent feature $\hat{\bm{U}}^\ast = (\hat u^\ast_1,\ldots, \hat u^\ast_b)^\top$ reside in $\mathcal{X}^\bot$, and let features $u^\ast_i$ be independent realizations of some distribution with mean zero and variance $\sigma_U^2$. Further assume that features ${\bm{X}}^\ast = ( x^\ast_1, \ldots,  x^\ast_b)^\top$ are realizations of some other distribution with mean zero and variance $\sigma_X^2$. Then $\sigma_E^2$ is $\mathcal{O}(1/b)$.
\end{lemma}
A corresponding proof can be found in the Appendix. This result is also confirmed in our simulations, showing that the additional prediction error induced by the orthogonalization decreases at a linear rate. While this is usually considered a good rate of convergence, note that this rate is w.r.t.~the {batch size} $b$, which does not grow in practice but is usually fixed to a moderately small number (e.g., $b=32$). In other words, the linear decrease in the additional prediction error has only limited benefits and might even be irrelevant in practice if the test set is small and hence with no option to increase $b$.
\begin{figure}
    \centering
    \includegraphics[width=\columnwidth]{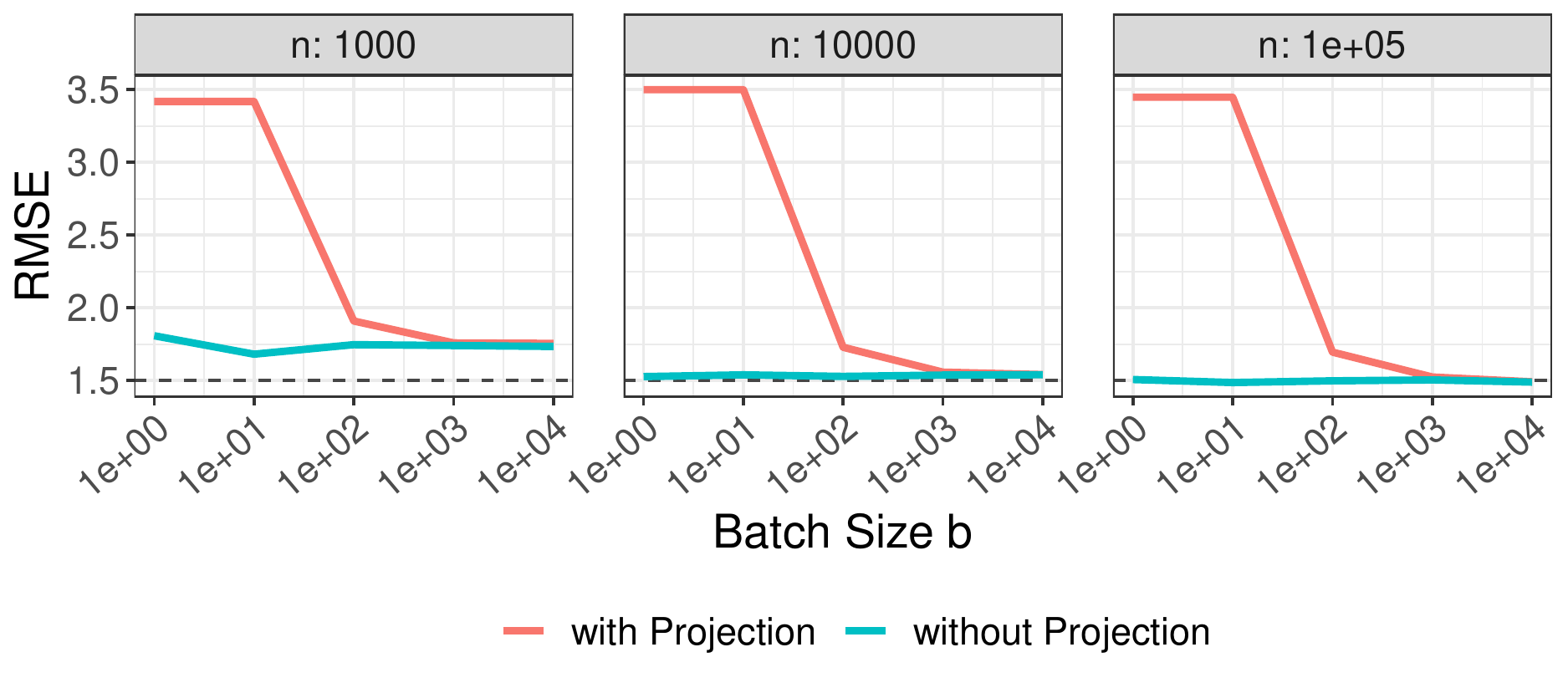}
    \caption{Root mean squared error (RMSE) of predictions for different training set sizes (facets) and different batch sizes (x-axis) when activating and deactivating the orthogonal projection. Results show that deactivation always leads to an optimal error which is also achieved in the limit as $b\to\infty$. For increasing $n$ both approaches further converge to the Bayes error (dashed black line).}
    \label{fig:ozbias}
\end{figure}
Fortunately, this problem can be easily solved. Define the estimated network weights for the structured and unstructured model part as $\hat{\bm{\beta}}$ and $\hat{\bm{\gamma}}$. Then the mean prediction on the test data is $$\bm{\eta}^\ast = \bm{X}^\ast \hat{\bm{\beta}} + \mathcal{P}_{X^\ast}^\bot \hat{\bm{U}}^\ast \hat{\bm{\gamma}} = \bm{X}^\ast \hat{\bm{\beta}} + \hat{\bm{U}}^\ast \hat{\bm{\gamma}} - \mathcal{P}_{X^\ast} \hat{\bm{U}}^\ast \hat{\bm{\gamma}},$$ where the last term $\mathcal{P}_{X^\ast} \hat{\bm{U}}^\ast \hat{\bm{\gamma}} \to \bm{0}$ for $b \to \infty$ as the SSN has been trained to learn features $\hat{\bm{U}}$ with entries (rows) residing in $\mathcal{X}^\bot$. Thus, predictions of an SSN with \acrold~in expectation (and in the limit) are given by $\bm{X}^\ast \hat{\bm{\beta}} + \hat{\bm{U}}^\ast \hat{\bm{\gamma}}$. In other words, if the \acrold~has successfully learned to generate features $\hat{\bm{U}}$ orthogonal to the space $\mathcal{X}$, the orthogonalization cell can be deactivated at test time (also confirmed in practice as shown in Figure~\ref{fig:ozbias}).

While the previous result implies that inferior prediction performance due to test set size dependency can be solved straightforwardly, it does not solve the other issues of \acrold.



\section{Post-Hoc Orthogonalization (PHO)}

To overcome the remaining previously mentioned problems, we propose the procedure described in Algorithm~\ref{alg:pomodoro} and dubbed \acr~(Post-Hoc Orthogonalization).
\begin{algorithm}[h]
   \caption{\acr}
   \label{alg:pomodoro}
\begin{algorithmic}
   \STATE {\bfseries Input:} Data set $\mathcal{D}$; unconstrained SSN
   \STATE 1.~Train SSN on $\mathcal{D}$
   \STATE 2.~Replace $\bm{\hat U}\bm{\hat{\gamma}}$ with $\mathcal{P}^\bot_X \bm{\hat U} \bm{\hat\gamma}$ in the last layer
   \STATE 3.~Set $\bm{\hat\beta} = \bm{\hat\beta} + \bm{X}^\dagger \bm{\hat U} \bm{\hat\gamma}$ with $\bm{X}^\dagger$ being the Moore-Penrose pseudoinverse of $\bm{X}$
   \STATE {\bfseries Output:} Return adapted SSN
\end{algorithmic}
\end{algorithm}

The underlying idea of \acr~is that all constraints proposed in \citet{sddr} only relate to the identifiability of $\bm{\beta}$, but not to the identifiability of the additive predictors $\bm{\eta}$. In other words, the unconstrained model is able to find the same $\bm{\eta}$ as the constrained model, implying that the optimal amount of structured effects needs to be contained in $\bm{\eta}$ in some form. Otherwise, the optimization routine could just change $\hat{\bm{\beta}}$ and thereby improve the predictions $\hat{\bm{\eta}}$. Moreover, the unconstrained model is not only theoretically able to learn the same effects as the one with \acrold~but will have more degrees of freedom in the optimization in practice. It is therefore not unlikely that it will find a similar or even better solution compared to the constrained model. The adaptions of fitted effects of the unconstrained SSN in steps 2 and 3 of Algorithm~\ref{alg:pomodoro} can then be justified as follows. Given the final (not meaningfully interpretable) model output $\bm{\hat\eta}$ we have
\begin{equation}\label{eq:neworthog}
\begin{aligned}
    \bm{\hat\eta} &= \bm{X}\bm{\hat\beta} + \bm{\hat U} \bm{\hat \gamma} = \bm{X}\bm{\widetilde\beta} + \bm{\widetilde U} \bm{\hat \gamma},\\
   \text{with } \bm{\widetilde\beta} &= \bm{\hat\beta} + \bm{X}^\dagger \bm{\hat U} \bm{\hat\gamma} \text{\,\, and \,\,} \bm{\widetilde U} = \mathcal{P}^\bot_X \bm{\hat U},
\end{aligned}
\end{equation}
where $\bm{X}^\dagger$ is the Moore-Penrose pseudoinverse of $\bm{X}$. The reformulation stems from the fact that we can split any predictor into 
\begin{equation} \label{eq:neworthog_long}
\begin{aligned}
\bm{\hat\eta}_k &= \bm{X}\bm{\hat\beta} + \bm{\hat U} \bm{\hat \gamma}\\
&= \bm{X}\bm{\hat\beta} + (\mathcal{P}_X \bm{\hat U} + \mathcal{P}^\bot_X \bm{\hat U}) \bm{\hat \gamma}\\
&= \bm{X}\bm{\hat\beta} + \bm{X}\bm{X}^\dagger \bm{\hat U} \bm{\hat\gamma} + \mathcal{P}_X^\bot \bm{\hat U}\bm{\hat \gamma}\\
&= \bm{X}(\bm{\hat\beta} + \bm{X}^\dagger \bm{\hat U} \bm{\hat\gamma}) + \mathcal{P}^\bot_X \bm{\hat U} \bm{\hat \gamma}\\
&=: \bm{X}\bm{\widetilde\beta} + \bm{\widetilde U} \bm{\hat \gamma},\\
\end{aligned}
\end{equation}
where $\mathcal{P}_X$ is the projection matrix into the column space spanned by $\bm{X}$ and $\mathcal{P}^\bot_X$ its orthogonal complement. In other words, $\bm{\widetilde\beta}$ is defined such that all linear effects of $\bm{X}$ found in $\hat{\bm{U}}\hat{\bm{\gamma}}$ are moved to the structured model part. From \eqref{eq:neworthog} it follows that the adjusted structured effects are $\bm{\widetilde\beta}$ and the unstructured predictions are $\bm{\widetilde U} \bm{\hat \gamma}$. Notably, as also apparent from \eqref{eq:neworthog}, the prediction of the model is not affected by this reformulation. Similarly, other constraints required to identify a model can be applied post-hoc (e.g., mean-centering to identify the global bias term of the model). 

\subsection{Properties, Efficient Computation and Prediction} 

Why does using \acr~solve the aforementioned problems? First, training the SSN unconstrained removes the constrained-induced difficulties in optimizing the SSN. Second, while Limitation~\ref{lim:null1} still exists also for \acr, we now have the guarantee that this does not deteriorate the prediction performance (as \eqref{eq:neworthog} states that predictions are unchanged by \acr). Moreover, this finding also solves Limitation~\ref{lim:size} as no orthogonalization is required for predictions (and, in contrast to \acrold, not part of the network). As \acr~is applied on the final model output and does not require any form of mini-batch updates, computations in \eqref{eq:neworthog} can be applied on the whole data set and thereby solves Limitation~\ref{lim:null2}. 

A possible downside of \acr~is that $\mathcal{P}_X^\bot \in \mathbb{R}^{n\times n}$ is potentially very large and its computation or storage is not possible. An alternative and computationally more efficient algorithm to compute \acr~is the following Algorithm~\ref{alg:pomodoro2} using the partitioning of the data set $\mathcal{D}$ into batches $\mathcal{B}_1,\ldots,\mathcal{B}_M$ with sizes $b_1,\ldots,b_M$. This algorithm is an \textbf{exact} procedure and is inspired by distributed computations of linear models \citep[e.g.,][]{karr2005secure}, i.e., no information is lost by switching to a mini-batch routine in this case, allowing us to scale our method arbitrarily. 
\begin{algorithm}[h]
   \caption{Mini-Batch \acr}
   \label{alg:pomodoro2}
\begin{algorithmic}
   \STATE {\bfseries Input:} Data set $\mathcal{D}$; unconstrained SSN; batches $\mathcal{B}_1,\ldots,\mathcal{B}_M$
   \STATE 1. Train SSN on $\mathcal{D}$
   \STATE 2. Initialize $\bm{H} = \bm{0}_{p\times p}$; $\bm{s} = \bm{0}_{p\times 1}$ and compute summary statistics as follows:
   \FOR{$m = 1,\ldots,M$}
    \STATE a) Compute $\bm{\hat U}_{\mathcal{B}_m}$ ($\bm{\hat U}$ for batch $\mathcal{B}_m$)
   \STATE b) Compute and store $\bm{\hat{\zeta}}_m := \bm{\hat U}_{\mathcal{B}_m} \bm{\hat \gamma} \in \mathbb{R}^{b_m \times 1}$
   \STATE c) Compute $\bm{H} = \bm{H} + (\bm{X}_{\mathcal{B}_m}^\top \bm{X}_{\mathcal{B}_m})$
   \STATE d) Compute $\bm{s} = \bm{s} + \bm{X}_{\mathcal{B}_m}^\top \bm{\hat{\zeta}}_m$
   \ENDFOR 
   \STATE 3. Compute $\bm{\alpha} := \bm{H}^{-1} \bm{s}$
   \STATE 4. Set $\bm{\widetilde\beta} = \bm{\hat\beta} + \bm{\alpha}$
   \STATE 5. Compute $\bm{\eta}^{unstr} = \texttt{stack}((\bm{\hat{\zeta}}_m - \bm{X}_{\mathcal{B}_m} \bm{\alpha})_{m=1,\ldots,M})$
   \STATE {\bfseries Output:} $\bm{\widetilde\beta}$, $\bm{\eta}^{unstr}$
\end{algorithmic}
\end{algorithm}
A similar algorithm can be used to calculate the different contributions of structured and unstructured network parts on a new data set $\mathcal{D}^\ast$ (see Appendix~\ref{app:oos}) when computing predictions. The total storage required for Algorithm~\ref{alg:pomodoro2} is $\mathcal{O}(p^2 + n)$. As $p$ is typically relatively small in semi-structured model applications and therefore negligible, the relevant memory requirement is to store a vector of size $n$ (i.e., the same size as the outcome vector $\bm{y}$). The time complexity of Algorithm~\ref{alg:pomodoro2} is $\mathcal{O}(c + n \cdot p^2 + p^3)$ where $c$ is the total cost of a single forward pass of the SSN using all $M$ mini-batches. Again, if $p$ is small and since $c$ is usually a multiple of $n$, the time complexity is practically not different from the one of a single forward pass (in batches) over the whole data set.

Note that it is straightforward to extend the presented algorithm to $K$ additive predictors $\bm{\eta}_k, k=1,\ldots,K$ as discussed in \citet{sddr} by simply iterating over $k$ and applying Algorithm~\ref{alg:pomodoro2} to all $K$ predictors.

\subsection{Semi-Structured Importance Measures}

As the last line of \eqref{eq:neworthog} creates two functional parts which are orthogonal to each other, the model decomposition follows the axioms proposed in the functional ANOVA approach \citep{hooker04}. We can thus derive importance measures post-hoc, relating to the variance explained by the different model parts (structured/unstructured) in the SSN. Various definitions are given in Appendix~\ref{app:imp}. 

\subsection{Relationship to other Methods}

The proposed approach opens up several links to other methods in the literature on model interpretability, which we want to briefly discuss in the following.

\textbf{Post-hoc Explanation Methods} \quad As mentioned in the previous subsection, the \acr~approach for SSNs is a functional ANOVA-type decomposition as introduced in \citet{hooker04}. Similar techniques have been applied to other machine learning models and have also been recently discussed for interaction effects and tree-based models \citep{lengerich2020purifying}. The SSN approach already specifies the hypothesis space using structural assumptions (for reasons discussed in the introduction) while other existing techniques assume no structure a priori and only impose the structure post-model fitting. 

\textbf{Feature Adjustment (Trend Adjustment)} \quad In statistical literature, feature (or trend) adjustment in linear models \citep[see, e.g.,][]{robinson1991some} follows a similar principle as our approach. By regressing the outcome on a feature or trend variable $\bm\chi$ that needs to be adjusted for, one can remove the influence of this variable before regressing the resulting linear model residuals (the remaining information in the outcome left after adjusting for $\chi$) on the actual set of features $\bm{X}$. In the special case of a linear model where the mean is parametrized by a semi-structured predictor, i.e., $\bm\eta = \bm{\eta}^{str} + \bm{\eta}^{unstr}$, our approach coincides with the feature adjustment approach by setting $\bm\chi = \hat{\bm\eta}^{unstr}$. For general parametric regression models, our approach can be seen as repeatedly applying the first step of feature detrending to (every of the $K$) additive predictor(s) separately.



\subsection{PHO with Splines} \label{sec:phogam}

Various approaches listed in Section~\ref{sec:rellit} use an SSN but define the structured model part $\bm{X}$ using a spline basis representation. This allows modeling structured non-linearities in $\bm{\eta}^{str}$ and thereby defining a combination of a DNN and a generalized additive model. In many of these applications, the structured part defined by the spline basis representation is, however, estimated using a penalized smoothing approach (i.e., using a quadratic difference penalty matrix for spline coefficients in $\bm{\beta}$). This leads to another limitation of \acrold~and, in this case, also of our \acr~approach.

\begin{limitation}[\textbf{Orthogonalization in penalized SSNs}] \label{lim:gam}
If the structured model part $\bm{\eta}^{str}$ in an SSN is estimated using a quadratic penalty matrix, the orthogonalization via $\mathcal{P}_X^\bot$ results in an ``over-orthogonalization'' by removing more contribution than necessary from $\bm{\eta}^{unstr}$.
\end{limitation}

We can overcome this restriction by accounting for the penalization when projecting the unstructured model part onto the space $\mathcal{X}$ as done in \eqref{eq:neworthog}. Limitation~\ref{lim:gam} can also be seen as some form of overfitting. When applying $\bm{X}^\dagger \hat{\bm{U}}\hat{\bm{\gamma}}$ in \eqref{eq:neworthog} without regularization, we are more likely to learn spline coefficients $\bm{\beta}$ that interpolate the data. Instead, given $\bm{X}$, the quadratic spline penalty $\bm{K}$ as well as the amount of penalization $\lambda \geq 0$, we can perform the corrected update
\begin{equation} \label{eq:adjpho}
    \widetilde{\bm{\beta}} = \hat{\bm{\beta}} + (\bm{X}^\top \bm{X} + \lambda \bm{K})^{-1} \bm{X}^\top \hat{\bm{U}}\hat{\bm{\gamma}}.
\end{equation}
While some of the related literature use a pre-defined smoothness $\lambda$ and the previous equation then directly yields the updated structured parameters, the projection in \eqref{eq:adjpho} can also be implemented by running a generalized additive model (GAM) routine with inputs $\bm{X}$, penalty matrix $\bm{K}$ and response $\hat{\bm{\eta}}^{unstr}$ to estimate $\lambda$ in a data-driven fashion. We call this approach PHOGAM in our experiments.

\subsection{Generalization Beyond Semi-structured Networks} \label{sec:general}

Lastly, we address one final limitation of \acrold~and present a generalization of the orthogonalization approach to a much larger class of networks beyond SSNs.
\begin{limitation}[\textbf{Model-class restriction of \acrold}] \label{lim:gen}
As \acrold~iteratively performs a projection into the space $\mathcal{X}^\bot$, $\mathcal{P}_X^\bot$ must be a non-random matrix and thus known prior to model training. 
\end{limitation}
The reason for $\mathcal{P}_X^\bot$ to be known a priori is that it would otherwise not be possible to calculate gradients for this projection. While this is not a limitation for the class of SSNs as the structured model part is always defined a priori by the modeler, Limitation~\ref{lim:gen} implies that the separation into structured and unstructured components can only be done if $\bm{X}$ is fixed. This, e.g., excludes cases such as neural additive models \citep[NAMs;][]{agarwal2021neural}, where additive components are learned features themselves and thus not known a priori. In contrast to \acrold, our PHO approach only applies the orthogonalization post-hoc and therefore allows its application to additive combinations with learnable inputs which only need to be fixed after training. This effectively generalizes the approach presented in \citet{sddr} to a much larger model class and allows the investigation of (non-)linearities in DNNs with learnable structured parts. 
%
%
\begin{figure}
    \centering
    \includegraphics[width=\columnwidth]{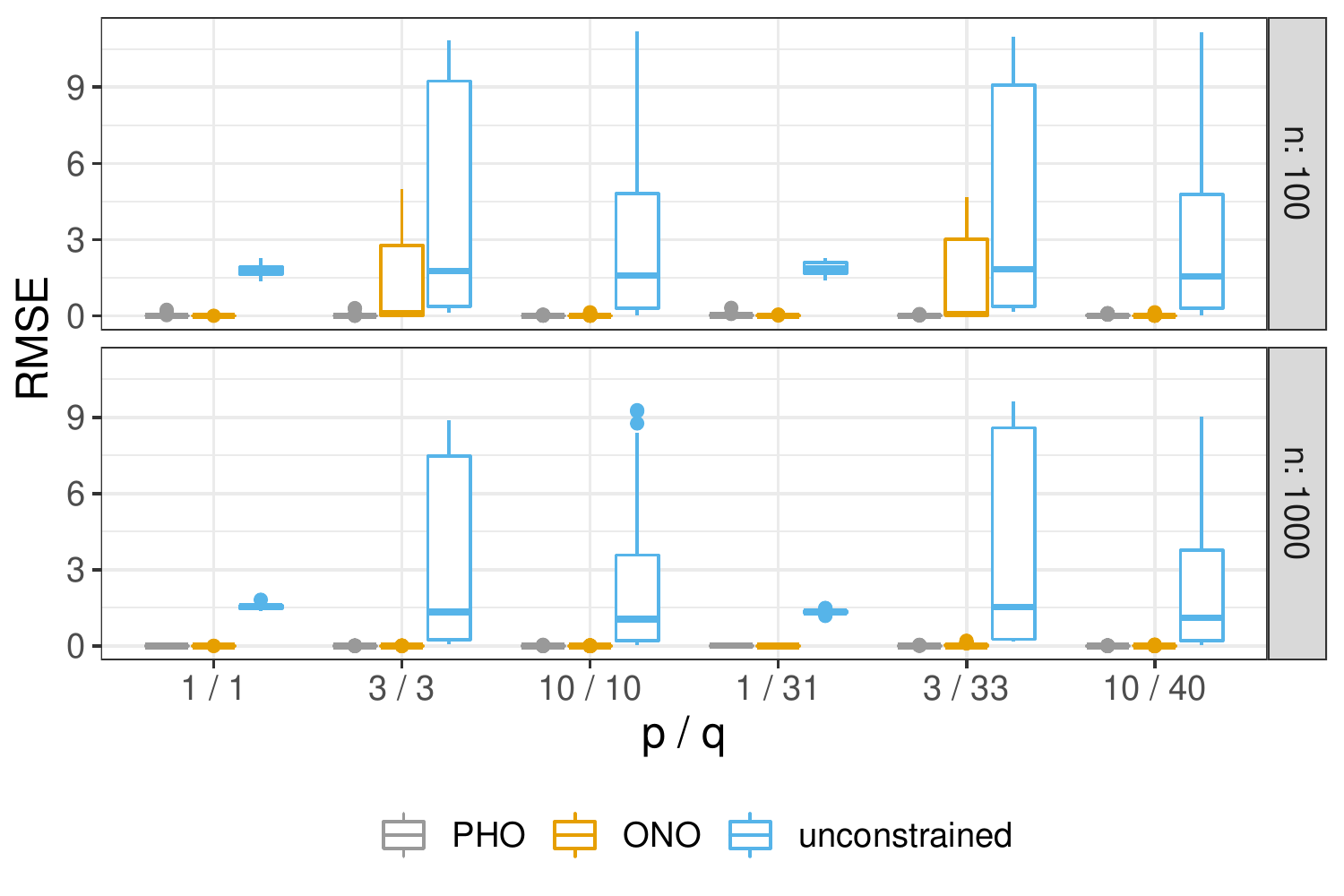}
    \caption{Linear coefficient estimation performance of different methods (colors) for different numbers of structured and unstructured features ($p$ / $q$, x-axis) and data sizes $n$ (rows).}
    \label{fig:linear_perf}
\end{figure}
\section{Numerical Experiments} \label{sec:numexp}

In the following, we investigate all derived properties of \acrold~as well as the performance of \acr~using simulated and real-world data. Further experiments and details such as optimization, computing environment, and hyperparameters can be found in Appendices~\ref{app:bench} and~\ref{app:addexp}. All code is made available on \hyperlink{https://github.com/davidruegamer/pho}{Github}. 

\subsection{Convergence of \acrold} \label{subsec:conv}

We first empirically investigate the convergence speed of \acrold~compared to \acr. As the convergence speed is influenced by the projection $\mathcal{P}_X$, we investigate different settings of structured features $p\in\{1,3,10\}$ as well as different numbers of observations $n\in\{100,1000\}$. 
We use a DNN with 2 hidden layers with 100 and 50 units and ReLU activation each followed by a dropout layer with a dropout probability of 20\%. A final layer learning the effects $\bm{\gamma}$ is defined by a linearly activated fully-connected layer. 
Results (see Figure~\ref{fig:convergence} in the Appendix) confirm our hypothesis, showing that additional iterations required by \acrold~compared to \acr~increase linearly with the number of features $p$, implying that the projection can slow down the training of an SSN notably. While this issue seems to be mitigated with more observations, \acrold~still requires a multiple of iterations of \acr~for $p=10$. 
\subsection{Semi-Structured Models with Linear Effects}
\begin{figure}
    \centering
    \includegraphics[width=\columnwidth]{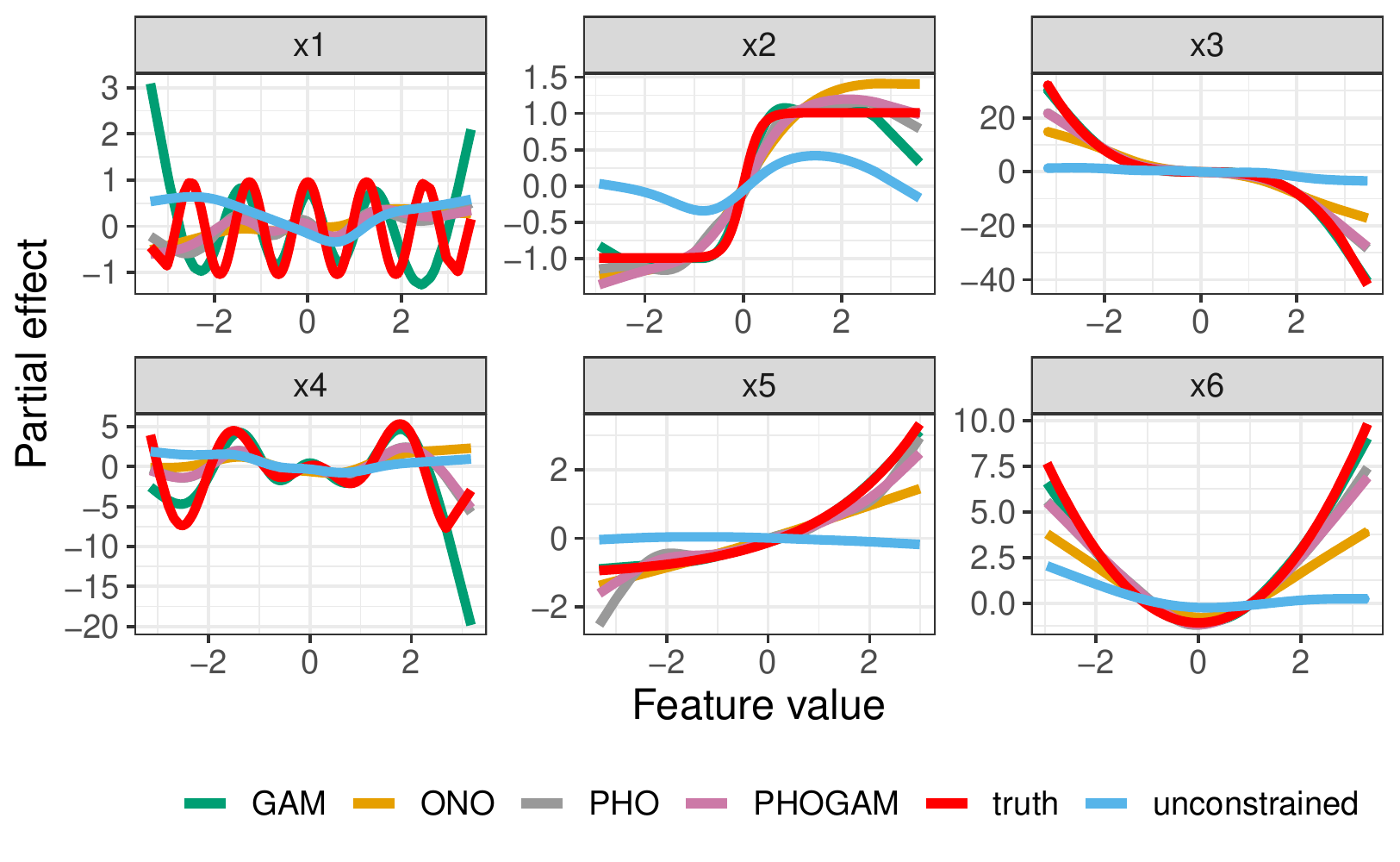}
    \caption{Visual comparison of different methods (colors) for six of the simulated non-linear functions (different facets; truth in red) visualized by the average curve over 20 repetitions (lines) for $n=1000$ and exact overlap between $\bm{X}$ and $\bm{Z}$.}
    \label{fig:splines}
\end{figure}
In our next numerical experiment, we investigate the estimation quality of structured effects $\bm{\beta}$ using~\acrold,~\acr, and an unconstrained model (Figure~\ref{fig:linear_perf}). We adopt the setup of the previous analysis to investigate different numbers of features $p$ and $q$ as well as the number of observations $n$. The deep neural network and optimization routine is defined as in Subsection~\ref{subsec:conv}. The estimation quality of the structured part is measured using the root mean squared error (RMSE) between the estimated and true model weights ($\bm{\beta}$) which are defined as equispaced values between $-2.5$ and $2.5$. The results are depicted in Figure~\ref{fig:linear_perf} showing that the unconstrained model is not able to find the right coefficients (the RMSE in most cases is larger than 1.5, which is the average absolute value of $\bm{\beta}$). In contrast, both \acrold~and \acr~perform well in general, but the optimization of \acrold~sometimes does not work perfectly, showing larger RMSEs in some cases.

\subsection{Semi-Structured Models with Non-Linear Effects} \label{subsec:nonlin}

We now investigate SSNs with non-linear structured effects based on a spline representation. We adopt the setup of the two previous subsections but define a non-linear structured relationship in the true data-generating process (i.e., the data-generating process is an additive model). The $p\in\{1,3,10\}$ non-linear functions along with further details are defined in Appendix~\ref{app:bench}. Six of the non-linear functions are visualized in Figure~\ref{fig:splines} together with estimated splines using \acrold, \acr, \acrgam, an unconstrained model and the gold-standard method for GAMs \citep{Wood.2017.book}. As Figure~\ref{fig:splines} suggests, the unconstrained model has difficulties properly estimating the non-linear effects while all other methods work similarly well. Quantitative results (Fig.~\ref{fig:nonlinear_perf} in Appendix~\ref{app:addexp1}) show that \acrgam~indeed improves over \acr~in some cases, suggesting that accounting for the model's penalization is beneficial. \acr, in turn, performs again better than \acrold~in most cases. All neural approaches exhibit a shrinkage effect towards zero, most likely caused by the implicit regularization of the first-order optimization routine. 

\subsection{Prediction Error of \acrold}

In a final simulation study, we investigate the additional prediction error of \acrold. For this, we create an SSN with~\acrold~and simulate random features $\bm{X}\in\mathbb{R}^{n\times 10}$ and $\bm{Z}\in\mathbb{R}^{n\times 20}$ from a standard random normal distribution, set $\bm{\beta}=\bm{0}$ and define the true non-linear data generating model as $\bm{y} = \sin(\bm{Z}_{[:,1]}) + \bm{Z}_{[:,2]}^2 + \bm{\varepsilon}, \bm{\varepsilon} \sim \mathcal{N}(\bm{0}, \bm{I})$. The latent features $\bm{U}$ are generated using a single hidden layer with 10 units, no bias, and ReLU activation. A final dense layer with one unit learns the last layer's weights $\bm{\gamma}$ using a linear activation. We vary $n\in\{1e4,1e5,1e6\}$ and train the network with an 80/20-split for training and validation data with early stopping the validation data with a patience of 50 iterations. We then evaluate the network on a separate test data set of size $n$ which we predict with a batch size $b\in\{1e0,1e1,1e2,1e3,1e4\}$. Finally, we evaluate the prediction performance using the RMSE both with activated and deactivated projection. We repeat this simulation 4 times and average the prediction errors across runs. The visual result from this simulation study was given in Section~\ref{sec:idenlim} in Figure~\ref{fig:ozbias}, confirming our theoretical understanding. We use the same data to also investigate the behavior of the additional prediction error $\sigma_E^2$ in terms of the batch size $b$ and find that numerically the rate is very close to $1/b$.

\subsection{Benchmark}
\begin{table}[!h]
\caption{Prediction performances (average MSE and its standard deviation in brackets) for different methods (rows) and data sets (columns) based on 10 train-test splits. The best method per data set is highlighted in bold.}
\label{tab:benchmark}
\begin{center}
\begin{footnotesize}
      \resizebox{\columnwidth}{!}{
\begin{tabular}{lrrrrrr}
\toprule
 & Airfoil & Concrete & Diabetes & Energy & ForestF & Yacht \\
\midrule
GAM & 119 (1.9) & 7.1 (0.53) & 140 (9.6) & 3.4 (0.36) & 1.5 (0.16) & 3.0 (0.43) \\ 
  DNN (large) & 22 (1.6) & 5.0 (0.66) & 63 (7.7) & \textbf{3.2} (0.39) & \textbf{1.4} (0.10) & 2.0 (0.44) \\ 
  DNN (small) & 23 (1.4) & 7.0 (1.40) & 58 (7.1) & 3.3 (0.31) & 1.4 (0.11) & 2.8 (0.54) \\ 
  ONO (large) & 113 (1.6) & 7.2 (0.55) & 120 (9.4) & 3.4 (0.36) & 1.5 (0.12) & 2.9 (0.43) \\ 
  ONO (small) & 115 (1.6) & 7.2 (0.56) & 130 (9.5) & 3.4 (0.36) & 1.5 (0.15) & 2.9 (0.43) \\ 
  PHO (large) & 14 (1.5) & \textbf{4.9} (0.80) & 57 (7.8) & 3.3 (0.36) & 1.5 (0.16) &  \textbf{1.9} (0.49) \\ 
  PHO (small) & \textbf{6} (0.4) & 6.3 (0.63) & \textbf{57} (7.2) & 3.3 (0.37) & 1.5 (0.13)  & 2.3 (0.56) \\ 
\bottomrule
\end{tabular}
}
\end{footnotesize}
\end{center}
\vskip -0.1in
\end{table}
The previous sections showed that \acr~and \acrold~yield the same structured effects while the first does not need any network adaptions before or during model training. We now compare both approaches in terms of prediction performance. As our approach is less restrictive during training and implies an easier optimization problem, we hypothesize that this will also result in better prediction performance. We, therefore, compare the two approaches (constrained vs. unconstrained) on different benchmark data sets from the UCI repository \citep{Dua.2019} using 10 different train-test splits. We report average MSE values and their standard deviation. In addition to the previously analyzed SSN models, we run the two obvious baselines, a model without an unstructured (neural network) predictor (i.e., a GAM) and a model without a structured (additive model) predictor (i.e., a DNN). For the models that include a DNN component, we choose between four pre-defined multi-layer perceptron architectures (see Appendix~\ref{app:bench} for more details).

Table~\ref{tab:benchmark} summarizes the results when choosing the best-performing hyperparameter set per method and data set. Further details on DNN architectures, hyperparameters, the full list of results, and details on benchmark data sets can be found in Section~\ref{app:bench} in the Appendix. Results indicate that \acr~works notably better in prediction compared to \acrold~and is almost on par with the DNN in cases where the deep unstructured model yields the best performance. In particular, for Airfoil, ONO (due to its limited flexibility) is effectively estimated as a GAM with almost zero DNN part, hence the similar performance to the GAM. Both DNN and \acr~perform similarly well but the additional inductive bias through the structural assumption of the SSN (\acr) allows further improvement over the DNN.
\begin{figure}
    \centering
    \includegraphics[width=0.9\columnwidth]{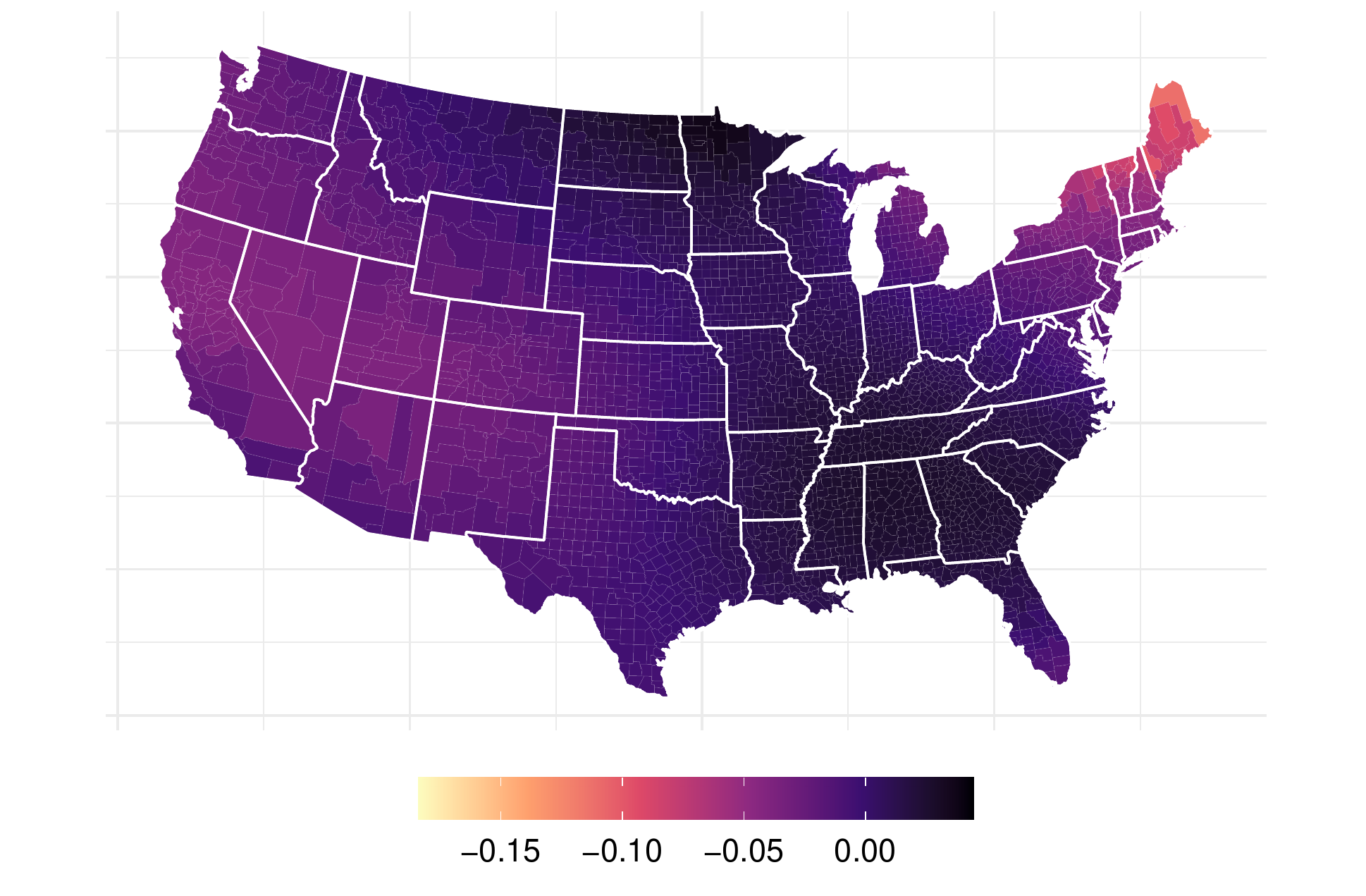}
    \caption{Spatial effect on the logarithmic prevalence of infection in the US estimated by PHOGAM (lower means infections are less likely).}
    \label{fig:spatial}
\end{figure}
\section{Real-World Applications} \label{sec:appl}

Finally, we demonstrate our approach by applying it to two large real-world data sets. 

\subsection{Covid-19 Analysis} \label{sec:covid}

We first analyze a recently published data set on global-scale spatially granular meta-data for coronavirus disease modeling \citep{wahltinez2022covid}. We subset the data to complete observations from the United States (US) and calculate the prevalence of infection per 1,000 people in the provided administration regions. We then model the logarithmic prevalence using an SSN including the average temperature, the relative humidity, the time (days since the start of the first outbreak), longitude, and latitude as well as the logarithmic population size as predictors. All features except the longitude and latitude are modeled as univariate splines whereas the geo-coordinates are modeled using a bivariate tensor-product spline to account for spatial interaction. We fit SSNs with \acrold, \acr, and \acrgam~and compare them with a semi-structured neural additive model \citep[NAM;][]{agarwal2021neural}. In contrast to the SSN approaches, the NAM learns each of the univariate non-linear effects as well as the bivariate non-linear effect using a feature-specific DNN. Details on network architectures and optimization routines can be found in Appendix~\ref{app:appl}. As \acr(GAM) can also work with learnable features, we apply \acr~also to the NAM (which we call PHONAM in the following) to demonstrate the generality of our approach discussed in Section~\ref{sec:general}.

\paragraph{Results} All models provide well-interpretable structured effects (see, e.g., Figure~\ref{fig:spatial} for a spatial effect visualization of \acrgam). When comparing the different approaches, we find similar patterns across models for all SSN approaches (Figure~\ref{fig:univ}, and Figure~\ref{fig:spatial2} in Appendix~\ref{app:appl}). In contrast to the pre-defined basis representation approaches (\acrold, \acr, \acrgam), the learned non-linear effects of NAM are less smooth as clearly visible in Figure~\ref{fig:univ} or show no trend (likely in cases when the optimization of the respective feature-specific network gets stuck in a local minimum). PHONAM is able to correct for missing smoothness, but can only learn as much additional feature information as present in NAM's deep predictor. Further effect comparisons can be found in Appendix~\ref{app:appl}.
\begin{figure}
    \centering
    \includegraphics[width=0.9\columnwidth]{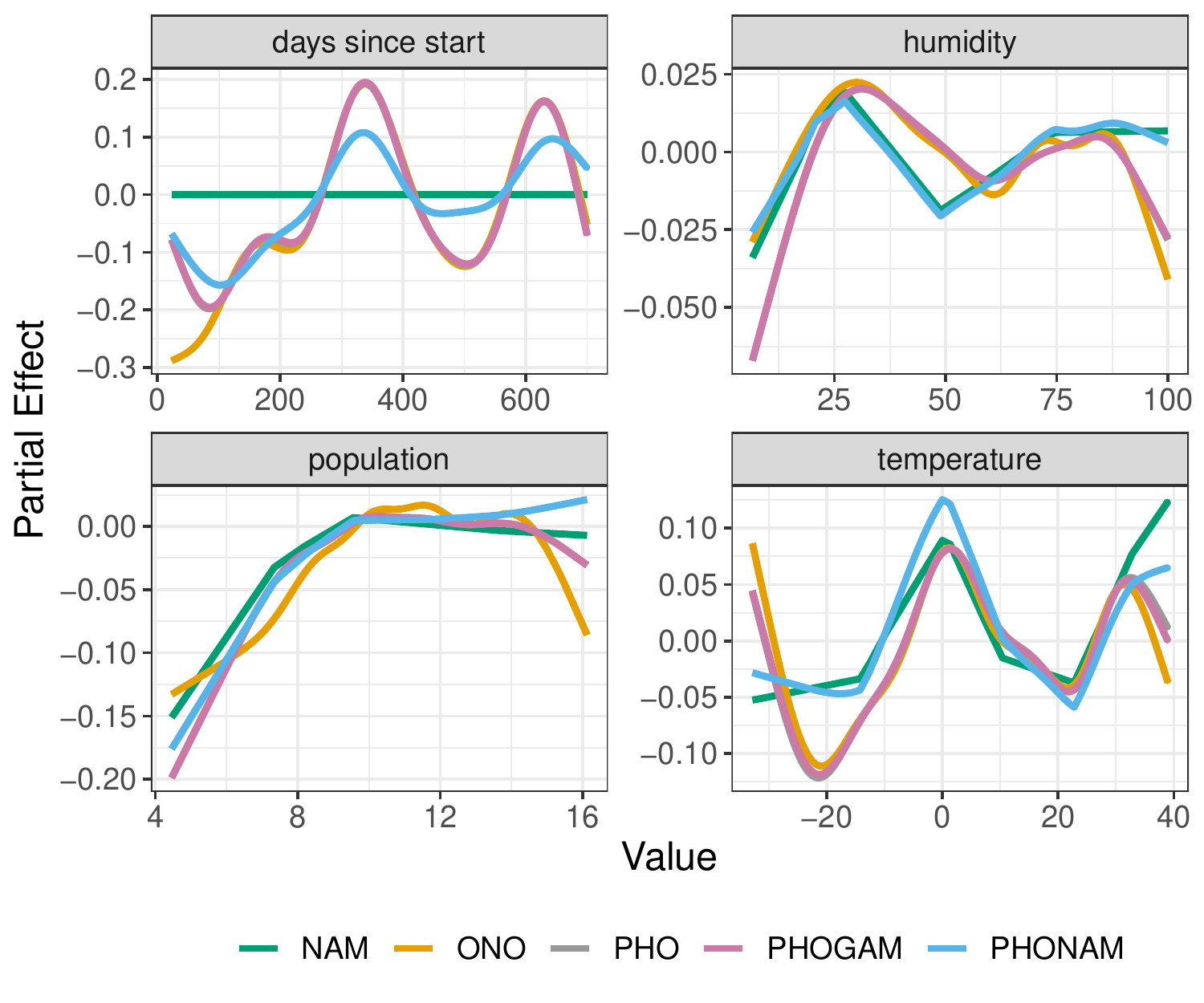}
    \caption{Estimated non-linear effects (mean-centered) for different features (facets) using different methods (colors).}
    \label{fig:univ}
\end{figure}
\subsection{Flight Delay Data}

To demonstrate the scalability of our approach, we further analyze the flight delay data set \citep{flights}. The data set contains detailed information about commercial flights within the US from October 1987 to April 2008, comprising approximately 120 million records. Data includes flight arrival and departure specifics, such as times, carrier codes, flight numbers, delays, cancellation details, and more. After filtering the data for non-canceled and non-diverted flights as well as routes and carriers that are present for all 22 years, we analyze the data with an SSN by modeling the year, the month, the day of the month, and the distance using splines, the scheduled departure and arrival time as cyclic splines (to enforce continuity for the transition between 11:59 pm and 0:00 am), as well as the origin, destination, their interaction, the day of the week and the carrier as factor effects (i.e., dummy-encoded binary features). The deep neural network also processing this information is defined by a 3-hidden layer ReLU network with 100 neurons each followed by 1 output neuron. 

\paragraph{Results}
\begin{figure}[!t]
    \centering
    \includegraphics[width=0.9\columnwidth]{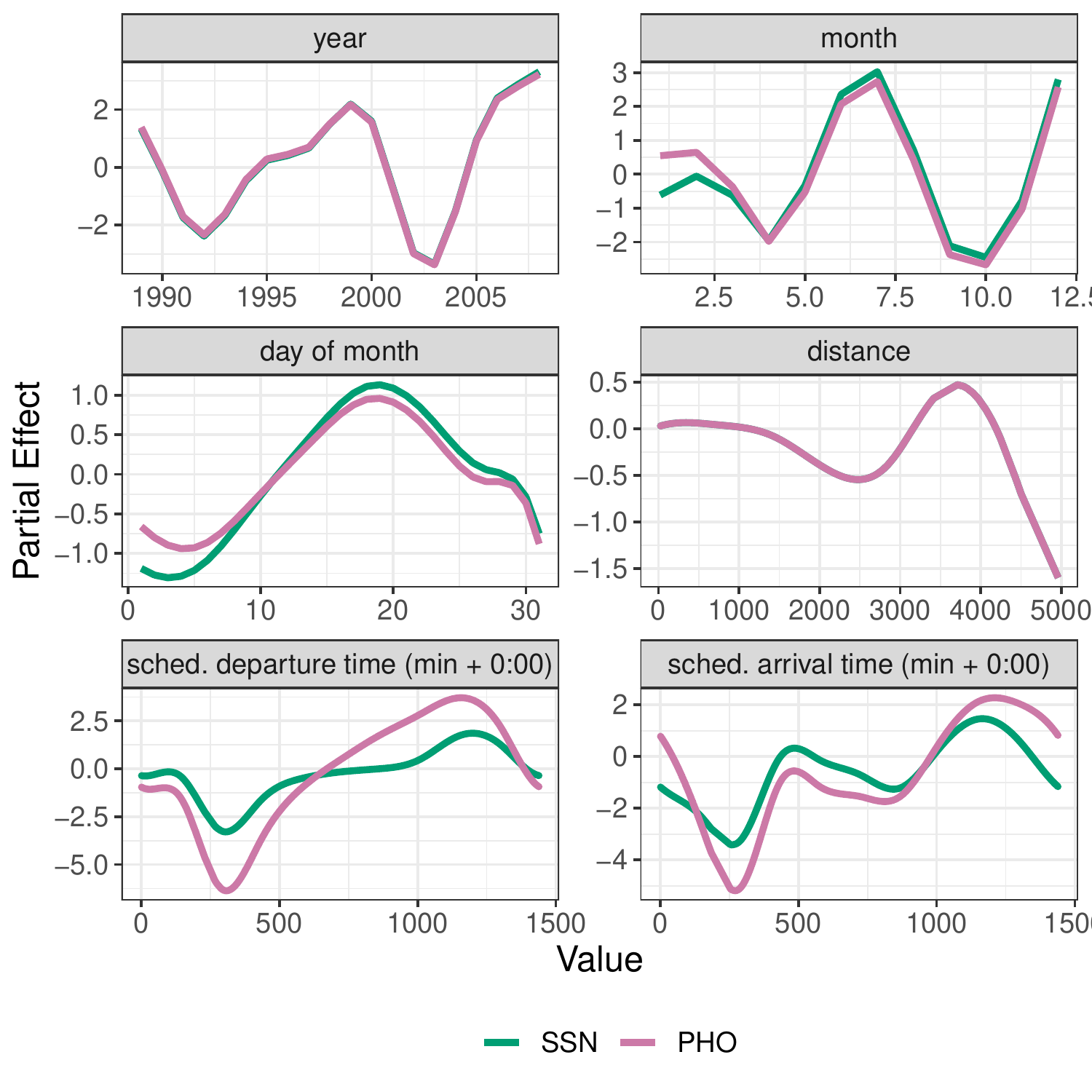}
    \caption{Partial effect (average arrival delay in minutes) of splines for all numerical features from both SSN (without adjusting for identifiability) and PHO.}
    \label{fig:flights}
\end{figure}
Upon inspecting the fitted model, it is apparent that the interaction effects modeled by the DNN do not substantially influence the prediction. This finding is supported by Figure~\ref{fig:flights}, which only displays minor adjustments after running PHO for the majority of variables. The exceptions are the scheduled arrival and departure times, which show more significant variation and larger effects when we extract their unidentified univariate information from the DNN component of the SSN. Overall, the estimated effects display a plausible pattern. For instance, flights scheduled early in the morning tend to be less delayed on average (indicating a negative partial effect). Conversely, flights during the summer and winter months typically experience more delays on average.

\section{Conclusion}

Semi-structured networks can be used to explicitly learn structured (non-)linear feature effects while jointly modeling other relationships using a deep network. The combination of structured and unstructured models can lead to identifiability issues, impeding the interpretability of the structured part. While this problem has been successfully addressed in previous work, we show that there exists an easy-to-implement, scalable, better-performing, and more general algorithm that theoretically circumvents the downsides of the previous solution and also excels in practice. 

\paragraph{Limitations}
While solving most of the limitations of previous work, our approach is not able to solve Limitation~\ref{lim:null1}. A possible solution could be to use \acrgam~together with a sparsity-inducing regularization. Another restriction of the approach is given when applied to learned features as presented in the previous section. If structured features learned by the network (e.g., by the feature-specific networks in NAMs) exhibit large correlation, an additional regularization is required to allow \acr~to extract additional linear information from the unstructured model part. 

\paragraph{Future Research}
Our approach opens up several future research directions. As mentioned in Section~\ref{sec:general}, \acr~-- in contrast to ONO -- can be applied to learned bases. Similar to the functional ANOVA approach \citep{hooker04}, this, e.g., also allows for a hierarchical decomposition post-model fitting, can be used to solve identifiability issues in NAMs, or to speed up networks that include structured assumptions.

\paragraph{Caveat}
We further want to emphasize that the given application in this paper only addresses a small subset of processes involved in the  spread of COVID-19 and should not be (the sole) basis for decision-making in the future. A more elaborate analysis could, e.g., impute missing information on the stringency index to explain some of the temporal dynamics now comprised in the time effect in our analysis. 

\section*{Acknowledgements}

We would like to thank the three anonymous reviewers for their comments that helped to further improve the paper. 

\bibliography{main}
\bibliographystyle{icml2023}

\clearpage

\appendix

\section{Proofs}

\subsection{Proof of Lemma~\ref{lemma:hyp}}

First, note that 
\begin{equation} \label{eq:hpho}
\mathcal{H}_{PHO} \equiv \mathcal{H},  
\end{equation}
where $\mathcal{H}$ is the hypothesis space of the unconstrained model. Eq.~\ref{eq:hpho} holds as \acr~first uses an unconstrained model to find the model parameters and given the model parameters $\hat{\bm \beta},\bm{\hat\gamma}$, \acr~just shifts explained variance from one to the other predictor by adjusting the weights, yielding the same model output, i.e.,
\begin{equation}
   \bm{\hat{\eta}} = \bm{X}\hat{\bm \beta} + \bm{\hat U} \bm{\hat\gamma} =  \bm{X}\widetilde{\bm \beta} + \bm{\widetilde U} \bm{\hat\gamma} = \bm{\hat \eta}.
\end{equation}
For ONO, it holds:
\begin{equation} \label{eq:hono}
\begin{split}
    &\mathcal{H}_{ONO} =\\ &\left\{ \bm{\hat \eta} = \bm{X}\hat{\bm \beta} + \bm{\hat U} \bm{\hat\gamma} : \hat{\bm \beta} \in \mathbb{R}^p, \hat{\bm{\gamma}} \in \mathbb{R}^q, \bm{\hat U} \in {\mathcal{X}^\bot}^{n \times q}  \right\},
\end{split}
\end{equation}
where $\mathcal{X}^\bot$ is the orthogonal complement of the column space 
$\mathcal{X}$ spanned by $\bm{X}$. This might already look like a hypothesis space restriction as for $\mathcal{H}$, we allow $\bm{\hat U}\in\mathbb{R}^{n \times q} \supset {\mathcal{X}^\bot}^{n \times q}$ while this is not the case for ONO. However, the following Corollary shows that this is, in fact, not restrictive.
\begin{corollary}
For every $\hat{\bm y}\in\mathcal{H}$ where $\bm{\hat U} \in \mathcal{X}^{n \times q}, n > p,$ there is always a parameter set $\bm{\breve \beta},\bm{\breve \gamma}$ with $\breve{\bm{U}} \in {\mathcal{X}^\bot}^{n \times q}$ that yields $\mathcal{H}_{ONO} \ni \bm{\breve{\eta}} = \bm{X}\bm{\breve \beta} + \breve{\bm{U}} \bm{\breve \gamma} = \bm{\hat \eta}$.    
\end{corollary}
The proof is, in turn, given by \eqref{eq:neworthog_long} when using $\bm{\hat y} = \bm{\hat \eta}_k$ yielding $\bm{\breve \beta} = \bm{\widetilde\beta}$, $\bm{\breve U} = \bm{\widetilde U}$, and $\bm{\breve \gamma} = \bm{\hat \gamma}$. As a consequence, the previous corollary also proves Lemma~\ref{lemma:hyp} since SSNs are a function of their additive predictor $\bm{\eta}$ only (and the same also holds for distributional regression models with $K$ predictors $\bm{\eta}_k, k=1,\ldots,K$).

\subsection{Proof of Lemma~\ref{lemma:error}}
In order to prove Lemma~\ref{lemma:error}, we make the following assumptions:
\begin{assumption}
Let $\hat u^\ast_1,\ldots, \hat u^\ast_b$ be independent realizations from a random variable $\mathfrak{U}$ with $\mathbb{E}(\mathfrak{U}) = 0$ and $\text{Var}(\mathfrak{U}) = \sigma_U^2$. Let $x^\ast_1,\ldots,x^\ast_b$ be independent realizations from a random variable $\mathfrak{X}$ with $\mathbb{E}(\mathfrak{X}) = 0$ and $\text{Var}(\mathfrak{X}) = \sigma_X^2$. 
\end{assumption}
Note that these assumptions are not very restrictive as 1) no distribution assumption is made, 2) the latent features $\hat{\bm{U}}$ are usually mean zero as the bias term is modeled in the structured model part in SSNs, 3) features $\bm{X}$ can always be mean-centered to fulfill the zero mean assumption, and 4) independence for data points are given per definition of the analysis setup (Section~\ref{sec:idenlim}). When the projection of the SSN works properly and $\mathfrak{U}$ and $\mathfrak{X}$ are orthogonal, i.e., $\mathbb{E}(\mathfrak{X} \mathfrak{U}) = 0$, then the inner product ${\bm{X}^\ast}^\top \hat{\bm{U}}^\ast = \sum_{i=1}^b x^\ast_i \hat u_i^\ast$ is the sum of $b$ independent variables with expectation zero. Therefore, by the central limit theorem, this inner product converges in distribution to a normal distribution with mean 0 and variance $\sigma_E^2 = \sigma^2_U \sigma^2_X b^{-1}$. Thus the variance of the inner product converges to $0$ with rate $\mathcal{O}(1/b)$ for $b\to\infty$. 

While this proves the limiting behavior of the variance induced by the projection, it does not discuss additional biases induced by it. However, as the mean of ${\bm{X}^\ast}^\top \hat{\bm{U}}^\ast$ is zero (also theoretically), the additional term $\mathcal{P}_{X^\ast} \hat{\bm{U}}^\ast \hat{\bm{\gamma}}$ in the projection also has mean zero, implying that the projection does not induce any additional bias but only additional variance yielding to the increase in RMSE. This variance is proportional to $\sigma_E^2$.

\section{Further Algorithms}

\subsection{Computing Out-of-sample Contributions} \label{app:oos}

Given a new data set $\mathcal{D}^\ast$ with structured features $\bm{X}^\ast$ and unstructured features $\bm{Z}^\ast$ and trained network on features $(\bm{X}, \bm{Z})$, the contributions of both network parts are given by 
\begin{align}
    \bm{\eta}_k^{str} &= \bm{X}^\ast \bm{\widetilde\beta},\\
    \bm{\eta}_k^{unstr} &= (\bm{\hat U}(\bm{Z}^\ast) - \bm{X}^\ast \bm{X}^\dagger \bm{\hat U}(\bm{Z})) \hat{\bm{\gamma}}, \label{eq:unstr}
\end{align}
where $\bm{\hat U}(\cdot)$ indicates the penultimate layer's features as a function of the unstructured network inputs. The contribution in \eqref{eq:unstr} follows from the fact that 
\begin{equation*}
   \bm{\eta}_k^{unstr} = \bm{X}^\ast \bm{\hat \beta} + \bm{\hat U}(\bm{Z}^\ast) \bm{\hat{\gamma}} - \bm{X}^\ast \bm{\widetilde\beta}
\end{equation*}
after orthogonalization.

\section{Semi-Structured Importance Measures} \label{app:imp}

Using the orthogonality property of predictors post-processed by PHO, we can define importance measures in the spirit of a functional ANOVA.

\begin{definition}{\textbf{Structured predictor importance}}\label{def:imp1}
The importance of a structured predictor in a post-hoc orthogonalized SSN can be characterized by $\text{EV}^{str} = {\text{Var}(\bm{X}\widetilde{\bm{\beta}})}/{\text{Var}(\hat{\bm{\eta}})}$.
\end{definition}

The explained variance $\text{EV}^{str}$ is a proportion $\in [0,1]$ and describes how much of the model's explanation can be related to the structured model part. Furthermore, a feature-based importance measure (also for SSNs beyond simple Gaussian mean regression) can be constructed as follows:

\begin{definition}{\textbf{Structured feature importance}}\label{def:imp2}
Let $\ell$ be the model's likelihood. Then a measure of importance for the $j$th structured predictor in $\bm{\eta}^{str}$ is $R^2_j := 1-(\ell(\hat{\bm{\eta}})/\ell(\hat{\bm{\eta}}_{-j}))$, where $\hat{\bm{\eta}}_{-j}$ is the predictor excluding the $j$th term in the structured predictor $\bm{\eta}^{str}$.
\end{definition}

The importance measure defined in Definition~\ref{def:imp2} is also known as McFadden's pseudo-$R^2$ \citep{mcfadden1973conditional}, yielding values between $0$ and $1$ (the higher, the more explained ``variance''). Both importance measures can also be further generalized to modeling setups with $K$ predictors in various ways. The structured predictor importance can, e.g., be calculated by considering the variance of only one of the $K$ structured predictors $\bm{\eta}^{str}_k$, yielding the importance of the structured model part for the $k$th parameter. For the structured feature importance, on the other hand, a meaningful definition would remove the $j$th feature from all predictors $\bm{\eta}_k, k=1,\ldots,K$, that include this feature.

\section{Simulation and Benchmark Details} \label{app:bench}

\subsection{Implementation}

To allow for a fair comparison between the methods, we use the package \texttt{deepregression} \citep{deepregression} that implements both GAMs, DNNs, and SSNs in TensorFlow \citep{Tensorflow} and R \citep{R}. This mitigates differences in performance due to different software implementations.

\subsection{Simulation Details}

For all simulations, we generate features using a standard normal distribution. The data-generating process of the simulation introduced in Subsection~\ref{subsec:nonlin} defines the 10 non-linear functions as follows:  
$f_0(x_0) = \cos(5 x_0)$, 
$f_1(x_1) = \tanh(3 x_1)$,
$f_2(x_2) = -(x_2^3)$,
$f_3(x_3) = -3x_3\cos(3 x_3 - 2)$,
$f_4(x_4) = \exp(0.5 x_4) - 1$,
$f_5(x_5) = x_5^2$,
$f_6(x_6) = \sin(x_6)\cos(x_6)$,
$f_7(x_7) = \sqrt{|x_7|}$,
$f_8(x_8) = \phi(x_8) - 1/8$ with standard normal density $\phi$, and
$f_9(x_9) = -x_9 \tanh(3 x_9) \sin(4 x_9)$.

\subsection{Benchmark Details}

\paragraph{Benchmark Data Sets}

Table~\ref{tab:further} gives an overview of the different data sets used in our benchmark.

\begin{table*}[]
\begin{tiny}
\begin{center}
\caption{Data set characteristics and references.} \label{tab:further}
\begin{tabular}{cccp{0.2cm}p{4cm}p{0.2cm}c}
Data set & \# Obs. & \# Feat. && Pre-processing && Reference \\ \hline 
Airfoil & 1503 & 5 && - && \citet{Dua.2019}  \\
Concrete   & 1030 & 8      &&  - &&  \citet{Yeh.1998} \\
Diabetes  & 442 & 10      &&     - &&  \citet{Efron.2004} \\
Energy  & 768 & 8          &&   - &&  \citet{Tsanas.2012} \\
ForestF   & 517 & 12 &&  logp1 transformation for \texttt{area}; numerical representation for \texttt{month} and \texttt{day} &&  \citet{Cortez.2007} \\
Yacht  & 308 & 6           &&   - &&  \citet{Ortigosa.2007,Dua.2019} \\ \hline
\end{tabular}
\end{center}
\end{tiny}
\end{table*}

\paragraph{Network Architecture and Optimization}

For all DNNs in our benchmark, we define architectures consisting of fully-connected layers, each with a $0.1$ dropout rate after every hidden layer. Hidden layers are defined using a ReLU activation, and the output layer uses no activation. The numbers of units of the four options are a) 200/1, b) 200/200/1, c) 20/1, and d) 20/20/1. All models are optimized using Adam \citep{kingma2014adam} with a learning rate of 0.001, and early stopping based on 10\% validation data with a patience of 50 iterations.

\paragraph{Table with all Results}

\begin{table*}[h]
\centering
          \resizebox{1.0\textwidth}{!}{
\begin{tabular}{rllllllllllll}
  \hline
 & GAM & ONO (a) & ONO (b) & ONO (c/d) & PHO (a) & PHO (b) & PHO (c) & PHO (d) & MLP (a) & MLP (b) & MLP (c) & MLP (d) \\ 
  \hline
Airfoil & 120 (1.9) & 110 (1.6) & 110 (1.8) & 110 (1.6) & 14 (1.5) & 3.4 (0.22) & 9.5 (1.2) & 5.9 (0.41) & 22 (1.6) & 3.3 (0.24) & 23 (1.4) & 5.8 (0.28) \\ 
  Concrete & 7.1 (0.53) & 7.2 (0.56) & 7.2 (0.55) & 7.2 (0.56) & 6 (0.58) & 4.9 (0.8) & 6.3 (0.63) & 7.6 (0.63) & 7.5 (1.5) & 5 (0.66) & 7.4 (0.87) & 7 (1.4) \\ 
  Diabetes & 140 (9.6) & 130 (9.5) & 120 (9.4) & 130 (9.5) & 57 (7.8) & 67 (9.5) & 57 (7.9) & 57 (7.2) & 63 (7.7) & 68 (8.5) & 62 (8.3) & 58 (7.1) \\ 
  Energy & 3.4 (0.36) & 3.4 (0.36) & 3.4 (0.36) & 3.4 (0.36) & 3.3 (0.36) & 3.4 (0.42) & 3.3 (0.37) & 3.3 (0.38) & 4.2 (0.44) & 3.2 (0.39) & 4.9 (0.56) & 3.3 (0.31) \\ 
  ForestF & 1.5 (0.16) & 1.5 (0.15) & 1.5 (0.12) & 1.5 (0.15) & 1.5 (0.16) & 1.7 (0.18) & 1.5 (0.13) & 1.5 (0.14) & 1.4 (0.1) & 1.6 (0.14) & 1.4 (0.11) & 1.4 (0.11) \\ 
  Yacht & 3 (0.43) & 2.9 (0.43) & 2.9 (0.43) & 2.9 (0.43) & 2.6 (0.42) & 1.9 (0.49) & 2.8 (0.46) & 2.3 (0.56) & 5 (0.54) & 2 (0.44) & 5 (0.61) & 2.8 (0.54) \\ 
   \hline
\end{tabular}
}
\end{table*}

\section{Additional Experiments and Results} \label{app:addexp}

\subsection{Convergence of \acrold}

The following Figure~\ref{fig:convergence} shows the additional iteration required by \acrold~compared to \acr~until the validation error is not improving anymore and early-stopping is triggered.

\begin{figure}
    \centering
    \includegraphics[width=0.81\columnwidth]{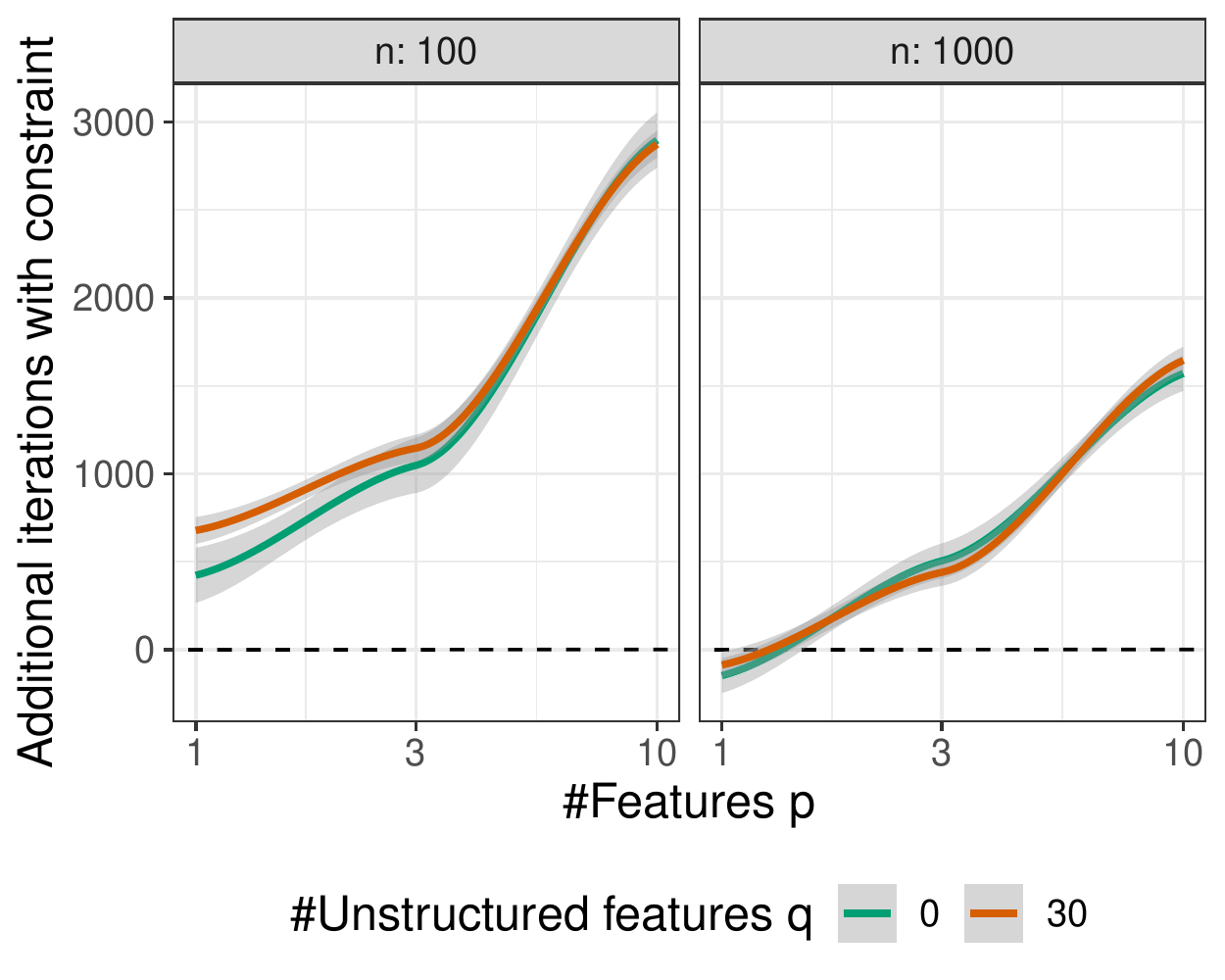}
    \caption{Difference in iterations when training the SSN with and without constraint (positive values indicate longer convergence with constraint) for different numbers of $p$ (x-axis), $q$ (colors), and data sizes $n$ (facets).}
    \label{fig:convergence}
\end{figure}

Results suggest a linear increase with $p$ whereas larger $n$ seems to mitigate the issue. However, as in general with more data fewer iterations are required, it is not directly clear whether this trend is indeed related to the projection. The size of $q$, i.e., the number of unstructured features, does not seem to play a role in the convergence speed. We, however, note that in general, many different factors can potentially influence the convergence and this simulation only shows that there is a downside in convergence speed which will get worse with increasing $p$.  

\subsection{Semi-Structured Models with Non-Linear Effects} \label{app:addexp1}

The results of Subsection~\ref{subsec:nonlin} are visualized in the following Figure~\ref{fig:nonlinear_perf}.

\begin{figure}
    \centering
    \includegraphics[width=\columnwidth]{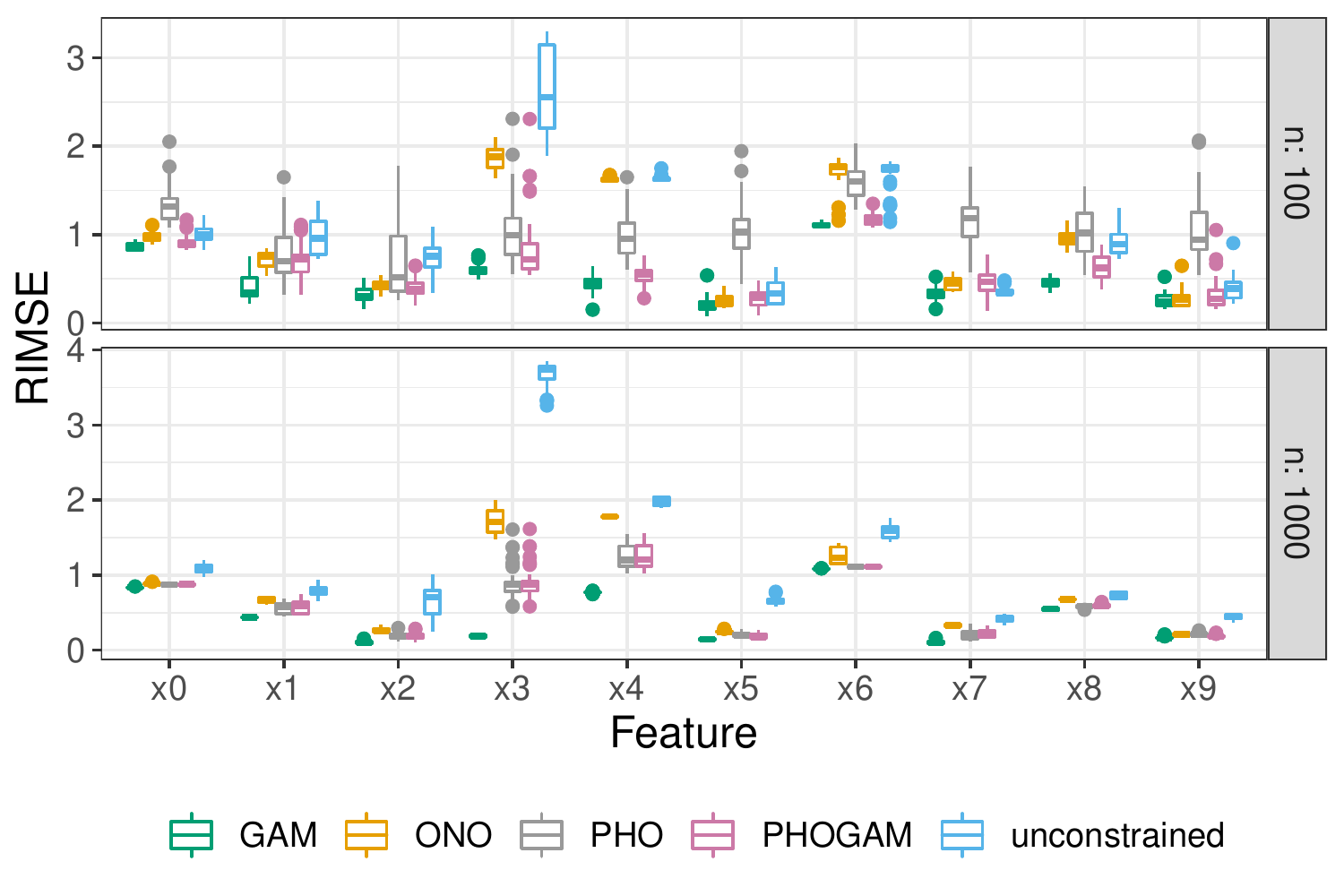}
    \caption{Spline estimation performance of different methods (colors) for different features (x-axis) and data sizes $n$ (rows).}
    \label{fig:nonlinear_perf}
\end{figure}
As for linear effects, we see that the unconstrained model is not able to correctly estimate the splines, yielding large RMSE values in most settings. As a gold standard, the GAM model defines a lower bound on the performance, which is often reached by the SSNs. While there is no clear trend or ranking between ONO, PHO, and PHOGAM, larger RMSE values are more often obtained using ONO. In various cases, PHOGAM further shows better performance than PHO, suggesting an improvement when accounting for the model's penalization.

\section{Further Details on the Application} \label{app:appl}

We now give further details on the application presented in Section~\ref{sec:appl}

\paragraph{Structured Model}
We model the time, temperature, humidity, and logarithmic population using a penalized thin plate regression spline with 9 basis functions and difference penalty following \citet{wood2003thin, Wood.2017.book}. The bivariate effect of longitude and latitude is modeled using a tensor-product spline using thin plate bases with a total of 25 basis functions. Penalization is again enforced through a difference penalty. 

\paragraph{Deep Neural Network}
For the DNNs for \acrold and \acr~we use 2 hidden layers each with 100 units, ReLU activation followed by a dropout layer with 20\% dropout rate. A final linear fully-connected layer with one unit is used to learn $\bm{\gamma}$. For the DNN of NAM we use a different DNN as we found that the otherwise either the feature-specific DNNs or the DNN for the unstructured part learns a relationship with the response but not both simultaneously. We, therefore, specify the DNN with two hidden layers each with 20 hidden units and tanh activation as well as a subsequent dropout layer with 20\% dropout rate. The feature networks are defined as in the software accompanying \citet{agarwal2021neural} using an activation layer (here with ReLU activation) and 64 units, followed by two hidden layers with 64 and 32 units (also activated using ReLU). The final layer is again specified using a linearly activated layer with 1 unit. For NAM's feature-specific DNN for the bivariate effect, we use the same architecture as for the other feature-specific DNNs for both latitude and longitude, but add another dense layer with 5 units and linear activation for both variables. We then calculate the row-wise tensor product of those two times 5 features, yielding the new DNN-learned feature basis of dimension 25 (analogous to the 25 basis function in the SSNs' structured predictors), which is finally combined with one dense layer with 1 hidden unit and linear activation.

\paragraph{Optimization and Evaluation} All networks are optimized using Adam \citep{kingma2014adam} with a learning rate of 0.001, a batch size of 1024 (chosen relatively large due to the size of the data set), 100 epochs, a validation split of 0.1, and a patience of 3 for early stopping. For NAM, we chose 150 epochs and a patience of 15 to account for the more erratic training curves.

\paragraph{Results}

\begin{figure*}[!h]
    \centering
    \includegraphics[width=0.8\textwidth]{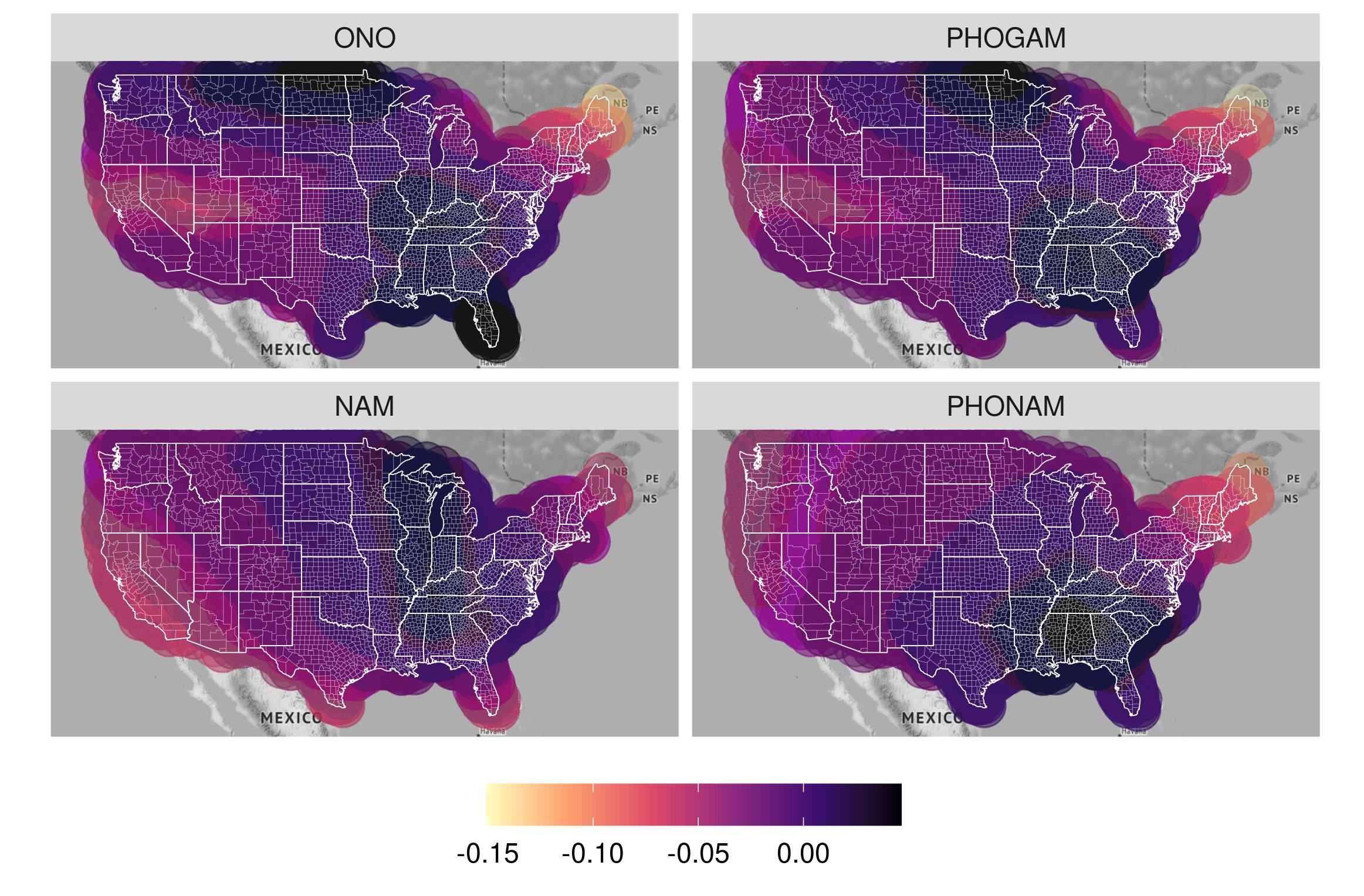}
    \caption{Estimated spatial effect of four of the methods (different facets).}
    \label{fig:spatial2}
\end{figure*}

Figure~\ref{fig:spatial2} shows the comparison between the estimated spatial effects of \acrold, \acrgam~(which is practically identical with \acr), NAM, and PHONAM. \acrold~and \acrgam~show very similar spatial effects whereas the NAM estimates a much more concentrated positive effect in the northeast of the US. The PHONAM variant corrects this effect and shows positive effects mainly in the south of the US.

\end{document}